\pdfoutput=1
\documentclass[11pt]{article}

\usepackage[final]{acl}

\usepackage{times}
\usepackage{latexsym}
\usepackage[T1]{fontenc}
\usepackage[utf8]{inputenc}
\usepackage{microtype}
\usepackage{inconsolata}
\usepackage{graphicx}
\usepackage{float}
\newcommand{\multirunstd}[1]{$_{\textcolor{gray}{\pm#1}}$}

\usepackage{amsmath}
\usepackage{amssymb}
\usepackage{booktabs}
\usepackage{tcolorbox}
\usepackage{xcolor}
\usepackage{algorithm}
\usepackage{algorithmic}
\usepackage{colortbl}
\usepackage{multirow}
\usepackage{pifont}
\usepackage{enumitem}
\usepackage{comment}
\usepackage{dblfloatfix}

\definecolor{icfblue}{RGB}{230,242,255}
\definecolor{icfborder}{RGB}{100,149,237}
\definecolor{mygrey}{gray}{0.9}
\newcommand{\thickhline}{\noalign{\hrule height 1pt}}

\title{Decomposition-Guided Diffusion Language Models for Inertial Confinement Fusion Prediction}

\author{
  Xiang Zhang\textsuperscript{1},
  Varchas Gopalaswamy\textsuperscript{2},
  Rahman Ejaz\textsuperscript{2},
  Riccardo Betti\textsuperscript{2},
  Dongfang Liu\textsuperscript{1}$^{\dagger}$
\\
  \textsuperscript{1}Purdue University,
  \textsuperscript{2}University of Rochester
\\
  $^{\dagger}$Corresponding author
}

\begin{document}
\maketitle

\begin{abstract}
Inertial confinement fusion (ICF) is a leading pathway toward clean energy, but each shot at the National Ignition Facility costs on the order of \$1M, making accurate AI surrogates a high-value target. We study \emph{exogenous-driven} ICF waveform prediction, where a 512-step neutron-rate diagnostic must be inferred directly from a laser pulse and target design parameters, with no historical response observed. The regime stresses standard time-series predictors with temporal sparsity (picosecond peak in a nanosecond window), input-output scale mismatch (${<}300$ real shots), and peak sensitivity (picosecond timing). We propose \textbf{ICF-DLM}, to our knowledge the first LM-based ICF predictor, combining (i) a physics-typed decomposition into yield $Y_{DT}$, peak timing $t_{\text{peak}}$, and local waveform $\mathbf{w}_{\text{local}}$; (ii) bidirectional denoising that defers commitment to peak location; and (iii) a physics-driven PPO reward re-injecting metric structure across numeric tokens. On \textbf{ICFBench} (50K simulations + 232 experimental shots), \textsc{ICF-DLM} cuts peak-timing error from $11.6$ to $9.2$ steps over a matched autoregressive LLaMA-3-8B and outperforms classical sequence models and LLM-based time-series predictors. Beyond ICF, the recipe shows potential to address science domains with low data and sparse events.
\end{abstract}

\section{Introduction}
\label{sec:intro}
\begin{figure}[h]
 \vspace{2mm}
    \centering
    \includegraphics[width=\columnwidth]{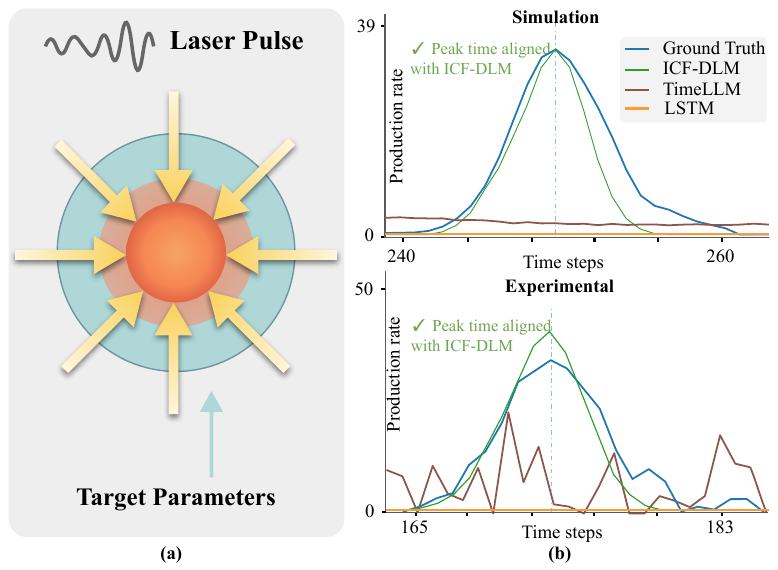}
    \vspace{-1.5em}
    \caption{(a) Exogenous-driven ICF waveform prediction maps a laser pulse and target parameters to a 512-step neutron-rate diagnostic. (b) \textsc{ICF-DLM} outputs on simulation (top) and experimental (bottom) data.}
    \vspace{-2em}
    \label{fig:teaser}
\end{figure}

Inertial confinement fusion (ICF) is a leading pathway toward clean energy \cite{nuckolls1972laser,lindl1995icf,Betti2016,Craxton2015,abushawareb2024nif}, but physical experiments are extraordinarily expensive. A single shot at the National Ignition Facility \cite{Radha2016,humbird2021cognitive} costs on the order of \textit{\$1 million}, with only a few feasible per year. High-fidelity radiation-hydrodynamics simulations are cheaper but remain computationally costly and suffer from sim-to-real domain shift \cite{humbird2019transfer}. This motivates a question for AI for science \cite{carleo2019mlphys,song2025ds}. Can pretrained language models serve as fast and accurate ICF surrogates?

We frame ICF prediction as exogenous-driven waveform generation. Given a time-resolved laser pulse and a few target design parameters, we predict the 512-step neutron-rate diagnostic with no historical observation of the response (Fig.~\ref{fig:teaser}(a)). The regime exposes three properties that break standard predictors. \textbf{\textit{(P1) Temporal sparsity:}} neutron emission occupies a picosecond window inside a nanosecond sequence, so pointwise objectives reward modeling the baseline rather than the event. \textbf{\textit{(P2) Input-output scale mismatch:}} a few hundred inputs determine the 512-step output, with only $\sim$300 real shots to anchor the mapping. \textbf{\textit{(P3) Peak sensitivity:}} small timing shifts cascade into large $L_2$ errors, so committing early to the wrong peak dominates the loss.\\
\indent Naively casting this task as next-token prediction with an autoregressive LM or as continuous regression with a Transformer leaves the structure of (P1)--(P3) unexploited. Autoregressive decoding commits to peak placement before the surrounding waveform is generated; deterministic regressors collapse to conditional means and obscure the heavy-tailed yield distribution \cite{salinas2020deepar}; classical sequence models trained on $\mathcal{O}(10^2)$ shots overfit and fail to transfer from simulation \cite{Craxton2015}. Diffusion language models \cite{ho2020ddpm,austin2021structured,li2022diffusionlm,nie2025llada} instead offer bidirectional refinement and a generative formulation amenable to reinforcement learning over numeric tokens, two ingredients aligned with (P1)--(P3).\\
\indent We present \textsc{ICF-DLM}, to our knowledge the first LM-based predictor for ICF. It addresses the above challenges through coordinated design choices. \textbf{(i) Physics-typed decomposition:} Following how physicists read radiation-hydrodynamics outputs, we factorize the waveform into total yield $Y_{DT}$, peak timing $t_{\text{peak}}$, and a short local waveform $\mathbf{w}_{\text{local}}$ around the peak. Predicting these descriptors jointly, rather than emitting 512 unstructured steps, concentrates supervision on the picosecond emission window and shrinks the effective output dimensionality to a few dozen tokens, both important under sparse events and a small experimental set. \textbf{(ii) Bidirectional denoising:} An LLaDA-8B \cite{nie2025llada} backbone refines all three descriptors in parallel under global bidirectional attention. The peak location is never committed before the surrounding waveform is generated, so small early-step mistakes can be corrected later, which is critical when timing shifts dominate the loss. \textbf{(iii) Physics-driven reward:} A composite PPO \cite{schulman2017ppo} reward scores magnitude accuracy, peak alignment, and waveform fidelity at the descriptor level. This re-injects the metric structure discarded by token-level cross-entropy, which treats adjacent numeric tokens as fully distinct classes, and supplies a continuous signal at the peak where small errors propagate to large $L_2$ deviations.\\
\indent We further curate \textbf{ICFBench}, pairing 50{,}000 simulation samples with 232 quality-controlled experimental shots, the largest ICF prediction benchmark to date. \textsc{ICF-DLM} outperforms classical sequence models, LLM-based time-series predictors, and a matched autoregressive LLaMA-3-8B trained with the same SFT+PPO recipe (Fig.~\ref{fig:teaser}(b); \S\ref{sec:benchmark}), with ablations isolating each design choice (\S\ref{sec:ablation}). Beyond ICF, the recipe (bidirectional generation, physics-typed factorization, metric-aware reward) may
 transfer to LM-based scientific prediction in low-data sparse-event regimes.

\section{Related Work}
\subsection{AI for Inertial Confinement Fusion}
Data-driven surrogates have been widely pursued where experiments and simulations are costly \cite{reichstein2019deep,carleo2019mlphys,lecun2015deep}, including in ICF, where prior work maps laser pulse and target parameters to yield, hot-spot quantities, and implosion dynamics \cite{gopalaswamy2019tripled,ejaz2024deep,ejaz2024can,gopalaswamy2025automated,leesExperimentallyInferredFusion2021,cao2022predicting,liu2025diff}, mostly through deterministic point-estimate models \cite{safavian1991survey,breiman2001random,hearst1998support,ahmed2010empirical,hatfield2019sparse}. Stronger inductive biases come from physics-informed and multi-fidelity learning \cite{raissi2019physics,karniadakis2021physics,li2020fourier,sanchez2020learning,han2018solving,meng2020composite}, yet point estimates underrepresent the intrinsic variability of ICF, where small perturbations and diagnostic noise produce large yield deviations \cite{Goncharov2008,Radha2016,gopalaswamy2019tripled,gaffney2024data}. Transfer learning narrows the sim-to-real gap \cite{pan2009survey,humbird2019transfer,humbird2022transfer}, yet to our knowledge no prior work applies \emph{pretrained language models} to ICF. \textsc{ICF-DLM} fills this gap, framing exogenous-driven waveform prediction as conditional sequence generation under (P1)--(P3) and releasing a benchmark pairing 50K simulations with 232 experimental shots.

\subsection{Language Models for Time-Series and Numeric Prediction}
Pretrained language models adapt to numeric time-series in two lines. \emph{Reprogramming-based} approaches (TimeLLM \cite{jin2024timellm}, LPI-LLM \cite{xue2024lpillm}) project continuous patches into the LM embedding space \cite{zhou2024one}. \emph{Tokenization-based} approaches discretize values into the LM vocabulary so next-token prediction performs regression \cite{wang2025chattime,golkar2023xval}. Both rely on autoregressive forecasting with modality-matched inputs and outputs, an assumption ICF violates under (P2). We therefore replace AR decoding with bidirectional diffusion (P3) and pointwise cross-entropy with a physics-driven reward; the matched-AR ablation (\S\ref{sec:ablation}) isolates the gain.\\
\indent Beyond LM-based predictors, dedicated architectures formulate physical evolution as recurrent \cite{hochreiter1997lstm,cho2014gru,ranzato2015sequence}, Neural-ODE \cite{chen2018neural,ruthotto2019deep,greydanus2019hamiltonian,cranmer2020lagrangian}, Transformer-variant \cite{vaswani2017attention,zhou2021informer,wu2021autoformer,zhou2022fedformer,nie2023patchtst,liu2024itransformer,wang2024timemixer}, or state-space \cite{gu2022s4,gu2024mamba} prediction; their pointwise objectives \cite{lim2021temporal,sapankevych2009time,leguen2019dilate} and strict autoregressive ordering \cite{ranzato2015sequence,laban2026multiturn} compound with (P3) to produce poor peak placement \cite{Craxton2015}. Diffusion-based predictors (CSDI \cite{tashiro2021csdi}, TimeDiff \cite{shen2023timediff}, Diffusion-TS \cite{yuan2024diffusionts}) sidestep the AR-ordering issue through iterative denoising, but use regression backbones on modality-matched horizons. \textsc{ICF-DLM} differs from these diffusion predictors in four respects, namely an exogenous-driven setting in which input and output share no modality, a pretrained diffusion LM with discrete numeric tokens rather than a regression backbone, a physically factorized output rather than a flat waveform, and PPO with a composite physics reward rather than denoising-only losses.

\subsection{Diffusion Language Models}

Discrete diffusion language models \cite{austin2021structured,li2022diffusionlm,savinov2022sundae,lou2024sedd,sahoo2024mdlm,nie2025llada,jinginfodlm} iteratively denoise masked tokens with bidirectional context, in contrast to the left-to-right factorization of autoregressive LMs \cite{brown2020gpt3,achiam2023gpt,touvron2023llama,llama3herd2024,chowdhery2022palm}. LLaDA \cite{nie2025llada} matches strong autoregressive baselines on language tasks while offering parallel generation, global conditioning, and an iterative refinement schedule \cite{ho2020ddpm} that can revise earlier commitments as later tokens crystallize. We exploit this revision capability against (P3), committing to peak location only after the surrounding waveform has been refined. We are the first, to our knowledge, to combine a pretrained diffusion LM \cite{nie2025llada}, numeric tokenization \cite{wang2025chattime,golkar2023xval}, and physics-driven RL reward shaping for scientific time-series prediction.

\section{Methods}

\begin{figure}[!t]
\centering
\includegraphics[width=\columnwidth]{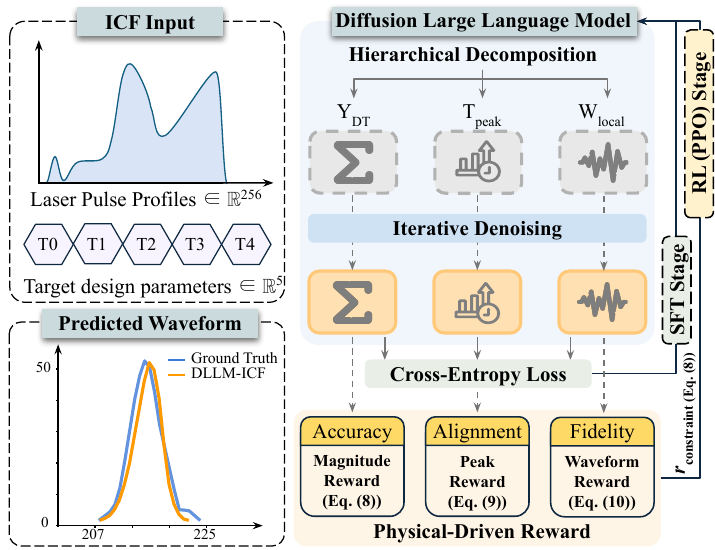}
\vspace{-1em}
\caption{\textsc{ICF-DLM} pipeline. The three physical descriptors $(Y_{DT}, t_{\text{peak}}, \mathbf{w}_{\text{local}})$ are denoised jointly under physics-driven PPO rewards for magnitude, peak timing, and waveform fidelity.}
\label{fig:overview}
\vspace{-1.5em}
\end{figure}

\textsc{ICF-DLM} (Fig.~\ref{fig:overview}) maps a laser pulse and target parameters to a 512-step neutron-rate waveform through three stages, namely input encoding, joint denoising of the physical descriptors $(Y_{DT}, t_{\text{peak}}, \mathbf{w}_{\text{local}})$ under physics-driven PPO rewards, and waveform reconstruction.

\subsection{Problem Setup}

\paragraph{Physical context.}
In ICF, high-power lasers compress a spherical deuterium-tritium (DT) target. As the fuel reaches extreme temperatures and densities, it initiates the fusion reaction:
\begin{equation}
\text{D} + \text{T} \rightarrow \text{n}(14.1\,\text{MeV}) + \alpha(3.5\,\text{MeV}).
\end{equation}
The experimental conditions are encoded by a composite input $\mathbf{x} \in \mathbb{R}^{261}$ formed by five \textbf{target parameters} $\vec{T} \in \mathbb{R}^5$ (e.g., outer radius, shell/ice thickness) and a \textbf{pulse shape} $\vec{P} \in \mathbb{R}^{256}$, the time-resolved laser power that drives the implosion.

\paragraph{Output waveform.}
The primary implosion diagnostic is the neutron rate $\dot{n}(t)$. We define the output $\mathbf{y} = [\dot{n}(t_1), \ldots, \dot{n}(t_T)] \in \mathbb{R}^T$ with $T{=}512$. The waveform has a sharp emission peak at the \textit{bang time} $t_{\text{peak}}$ and is negligible elsewhere, reflecting (P1). The total yield is the integral:
\begin{equation}
Y_{DT} = \int_{0}^{T} \dot{n}(t) \, dt.
\end{equation}

\paragraph{Problem formulation.}
We learn $f \mathpunct{:} \mathbf{x} \mapsto \mathbf{y}$ predicting the full neutron-rate waveform via a hierarchical physics-typed factorization that decomposes $\mathbf{y}$ into the yield ($Y_{DT}$), peak timing ($t_{\text{peak}}$), and local shape ($\mathbf{w}_{\text{local}}$).

\subsection{\textsc{ICF-DLM}}

\subsubsection{Diffusion Language Modeling}\label{subsec:dlm}

The descriptors $(Y_{DT}, t_{\text{peak}}, \mathbf{w}_{\text{local}})$ are emitted as discrete tokens. Continuous values in $[-0.5, 0.5]$ map to a vocabulary $\mathcal{V}_{\text{num}}$ of $V{=}10{,}000$ uniformly-spaced bins, and the structured prompt $\mathbf{x}{=}\mathrm{tok}(\vec{P}, \vec{T})$ and the masked output tokens form the model input.

LLaDA \cite{nie2025llada} couples a forward masking process with a reverse unmasking process. Given the clean descriptor tokens $\mathbf{z}_0$ (the tokenized $(Y_{DT}, t_{\text{peak}}, \mathbf{w}_{\text{local}})$) and a mask ratio $\beta \in [0,1]$, the forward process replaces a fraction $\beta$ of positions with a $[\textsc{m}]$ token. The reverse process parameterizes the masked positions through a bidirectional Transformer:
\begin{equation}
p_\theta(\mathbf{z}_0 \mid \mathbf{z}_\beta, \mathbf{x}) = \prod_{i \in \mathcal{M}} p_\theta\bigl(z_0^{(i)} \mid \mathbf{z}_\beta, \mathbf{x}\bigr),
\end{equation}
trained by the masked cross-entropy loss:
\begin{equation}
\mathcal{L}_{\text{SFT}} = -\, \mathbb{E}_{\beta \sim U(0,1)} \sum_{i \in \mathcal{M}} \log p_\theta\bigl(z_0^{(i)} \mid \mathbf{z}_\beta, \mathbf{x}\bigr).
\end{equation}
At inference we denoise over $S{=}128$ steps with confidence-based remasking and block size $128$ (ablated in \S\ref{sec:ablation}).

\subsubsection{Physics-Typed Hierarchical Decomposition}
\label{subsec:3_1}

Targeting (P1) and (P2), we decompose the prediction into three physical descriptors inspired by radiation-hydrodynamics (rad-hydro) workflows \cite{delettrez1987lilac,radha2005draco,igumenshchev2016threedimensional}. Physicists routinely characterize ICF responses using scalar energy descriptors (e.g., yield $Y_{DT}$), temporal landmarks (e.g., bang time $t_{\text{peak}}$), and local waveform structure ($\mathbf{w}_{\text{local}}$); see Appendix~\ref{appendix:physical_interpretation}. Rather than predicting $\mathbf{y} \in \mathbb{R}^T$ directly, we predict:
\begin{equation}
(Y_{DT},\, t_{\text{peak}},\, \mathbf{w}_{\text{local}}) = f(\mathbf{x}; \theta),
\end{equation}
where $Y_{DT}$ is the integrated total yield, $t_{\text{peak}}$ is the peak timing, and $\mathbf{w}_{\text{local}} \in \mathbb{R}^{T_{\text{local}}}$ is the local waveform around the peak. The DLM denoises all three jointly at each step, with each descriptor receiving an independent reward (\S\ref{subsec:reward}). The full waveform is reconstructed at inference by embedding $\hat{\mathbf{w}}_{\text{local}}$ at $\hat{t}_{\text{peak}}$, with the remaining steps filled by a near-zero baseline (Appendix~\ref{appendix:components}). The factorization concentrates supervision on the peak window and reduces effective output dimensionality from $T{=}512$ to $\sim$$20$ tokens, jointly addressing the sparsity and low-data conditions of (P1)--(P2). 
We impose the physical identity $\hat{Y}_{DT} = \sum_t \hat{y}_t$ at
reconstruction. Specifically, the predicted local waveform is rescaled by
$\hat{Y}_{DT} / \sum_t \hat{w}_{\text{local},t}$ before being embedded at
$\hat{t}_{\text{peak}}$. In this manner, the yield descriptor and the
waveform descriptor remain consistent by construction.

%We predict $\hat{Y}_{DT}$ and $\hat{\mathbf{w}}_{\text{local}}$ independently and do not enforce the physical identity $\hat{Y}_{DT} = \sum_t \hat{y}_t$ as a hard constraint; a consistency penalty is left to future work.

\subsubsection{Physics-Driven Reward Modeling}\label{subsec:reward}

Optimization is coarse-to-fine \cite{ouyang2022instructgpt}. We first apply supervised fine-tuning with $\mathcal{L}_{\text{SFT}}$ (\S\ref{subsec:dlm}) \cite{chung2022scaling,wei2022finetuned} to initialize the hierarchical output format ($Y_{DT} \to t_{\text{peak}} \to \mathbf{w}_{\text{local}}$), and then continue with PPO \cite{schulman2017ppo}, which optimizes a clipped surrogate of the reward gain:
\begin{equation}
\begin{split}
\mathcal{L}_{\text{PPO}}(\theta) = & -\mathbb{E}\bigl[\min(\rho_\theta \hat{A},\, \mathrm{clip}_\epsilon(\rho_\theta)\, \hat{A})\bigr] \\
& + \beta_{\text{KL}}\, \mathrm{KL}(\pi_\theta \,\Vert\, \pi_{\text{ref}}),
\end{split}
\end{equation}
where $\rho_\theta$ is the per-token importance ratio, $\hat{A}$ the advantage from the composite reward $r$, $\pi_{\text{ref}}$ the SFT-initialized policy, and $\mathrm{clip}_\epsilon(\cdot) = \mathrm{clip}(\cdot, 1{-}\epsilon, 1{+}\epsilon)$. Token-level cross-entropy is insufficient. The three descriptors carry very different numerical scales and error sensitivities, and cross-entropy treats adjacent numeric tokens as fully distinct classes, inducing a metric mismatch under discrete tokenization. We therefore define the composite reward:
\begin{equation}
r = r_{\text{mag}} + r_{\text{peak}} + r_{\text{wave}} + r_{\text{constraint}},
\end{equation}
which re-injects metric structure across numeric tokens, targeting (P1) and (P3).

\paragraph{Magnitude accuracy reward.}
Yield $Y_{DT}$ is a scalar spanning several orders of magnitude across shots. 
To produce a bounded, scale-invariant signal, we use a reciprocal of the 
relative yield error.
\begin{equation}
r_{\text{mag}}(\hat{Y}_{DT}, Y_{DT}^*) = \frac{1}{1 + \left|\frac{\hat{Y}_{DT} - Y_{DT}^*}{Y_{DT}^*}\right|}.
\end{equation}
A perfect prediction maps to $1.0$ and large errors approach $0$ asymptotically.

\paragraph{Peak alignment reward.}
Peak timing exhibits sharp local sensitivity, since small temporal offsets 
cause large waveform deviations. To provide gradient near the peak while 
preserving signal for large timing errors, we use an exponential decay.
\begin{equation}
r_{\text{peak}}(\hat{t}, t^*) = \exp\left(-|\hat{t} - t^*|\right).
\end{equation}
This addresses the peak-sensitivity challenge identified in the introduction.

\paragraph{Waveform fidelity reward.}
The squared waveform error grows rapidly with per-step deviation and can 
dominate the combined reward. We apply a logarithmic compression to keep 
its contribution balanced.
\begin{equation}
r_{\text{wave}}(\hat{\mathbf{w}}, \mathbf{w}^*) = -\log\left(1 + \|\hat{\mathbf{w}} - \mathbf{w}^*\|_2^2\right).
\end{equation}
This balances peak timing, magnitude, and waveform fidelity in the combined reward.
\paragraph{Physical constraint reward.}
We penalize physically implausible predictions, namely negative yields or peaks outside $[0, T]$:
\begin{equation}
r_{\text{constraint}} = \begin{cases}
-\lambda_c & \text{if } \hat{Y}_{DT} < 0 \text{ or } \hat{t} \notin [0, T] \\
0 & \text{otherwise.}
\end{cases}
\end{equation}
The total reward is clipped to a bounded range to prevent gradient explosion while preserving relative ranking \cite{schulman2017ppo}.

\section{Experiments}\label{sec:experiments}

\subsection{Experimental Setup}

\subsubsection{Dataset and Benchmark}
We introduce the \textbf{ICFBench} dataset, the first large-scale benchmark for ICF neutron yield time-series prediction. The dataset comprises (1) \textbf{Simulation data}: 50{,}000 samples generated from radiation-hydrodynamics codes, split into 40{,}000 / 5{,}000 / 5{,}000 for train / validation / test; and (2) \textbf{Experimental data}: 232 quality-controlled physical shots split into 182 training shots and a 50-shot held-out test set used for all physical-experimental results in this paper. Each sample consists of 5 scalar target parameters ($\vec{T}$) and a 256-dimensional pulse shape ($\vec{P}$), with target neutron rate ($\dot{n}$) time-series of 512 steps. All models predict the neutron rate from target design parameters. Full dataset statistics, the quality-control protocol, and the per-table evaluation ledger are in Appendix~\ref{appendix:dataset}.

\subsubsection{Training Setup}
{We build on LLaDA-8B~\cite{nie2025llada} with LoRA adapters for parameter-efficient tuning~\cite{hu2021lora,wang2023aprompt,han2024facing}. Training proceeds in two stages: supervised fine-tuning followed by PPO with the physics-driven rewards in \S\ref{subsec:reward}. All experiments use 8 NVIDIA H100 GPUs. More details are in Appendix~\ref{appendix:implementation}.}

\subsubsection{Baselines}
 We compare ICF-DLM against seven baselines covering classical deep time-series models (RNN~\cite{ranzato2015sequence}, LSTM~\cite{hochreiter1997lstm}, GRU~\cite{cho2014gru}, Transformer~\cite{vaswani2017attention}, FEDformer~\cite{zhou2022fedformer}), diffusion-based time-series models (Diffusion-TS~\cite{yuan2024diffusionts}), and LLM-based time-series models (TimeLLM~\cite{jin2024timellm}).

% \subsubsection{Baselines}
% We compare against three categories of methods:
% \begin{itemize}[left=1pt]
%     \item \textbf{Classical deep time-series models}: RNN~\cite{ranzato2015sequence}, LSTM~\cite{hochreiter1997lstm}, GRU~\cite{cho2014gru}, Transformer~\cite{vaswani2017attention}, FEDformer~\cite{zhou2022fedformer}.
%     \item \textbf{LLM-based time-series models}: TimeLLM~\cite{jin2024timellm}, LPI-LLM~\cite{xue2024lpillm}.
%     \todored{\item \textbf{Matched autoregressive ablation}: LLaMA-3-8B with the same numeric tokenization, prompt, LoRA configuration, and SFT+PPO pipeline as ICF-DLM but replacing the diffusion backbone with autoregressive generation, isolating the contribution of the diffusion paradigm.}
% \end{itemize}

\begin{table*}[h]
\centering
\caption{Quantitative results on simulation and physical-experimental datasets for \textbf{exogenous-driven ICF waveform prediction}. Method categories: $\dagger$ Classical Deep Time-Series Models; $\flat$ Diffusion-based Time-Series Models; $\ddagger$ LLM-based Time-Series Models; $\star$ Ours.}
\label{tab:results}
\vspace{-0.5em}
\fontsize{8}{10}\selectfont
\setlength\tabcolsep{4pt}
\renewcommand\arraystretch{1.1}
\begin{tabular}{l|cccc|cccc}
\hline\thickhline
\rowcolor{mygrey}
 & \multicolumn{4}{c|}{\textbf{Simulation}} & \multicolumn{4}{c}{\textbf{Physical Experimental}} \\
\rowcolor{mygrey}
\multirow{-2}{*}{\textbf{Method}}& \textbf{MSE}$\downarrow$ & \textbf{MAE}$\downarrow$ & \textbf{PTE}$\downarrow$ & \textbf{YE (\%)}$\downarrow$ & \textbf{MSE}$\downarrow$ & \textbf{MAE}$\downarrow$ & \textbf{PTE}$\downarrow$ & \textbf{YE (\%)}$\downarrow$ \\
\hline\hline
RNN$^{\dagger}$         & 0.401\multirunstd{0.002} & 0.550\multirunstd{0.003} & 13.3\multirunstd{0.58} & 35.5\multirunstd{0.4}  & 0.304\multirunstd{0.014} & 0.472\multirunstd{0.018} & 15.2\multirunstd{0.76} & 303.8\multirunstd{35} \\
LSTM$^{\dagger}$        & 0.362\multirunstd{0.018} & 0.513\multirunstd{0.022} & 10.7\multirunstd{2.52} & 34.8\multirunstd{4.4}  & 0.304\multirunstd{0.019} & 0.472\multirunstd{0.016} & 20.7\multirunstd{6.83} & 404.5\multirunstd{165} \\
GRU$^{\dagger}$         & 0.338\multirunstd{0.006} & 0.480\multirunstd{0.010} & 9.0\multirunstd{1.00}  & 30.9\multirunstd{3.2}  & 0.236\multirunstd{0.022} & 0.394\multirunstd{0.006} & 25.7\multirunstd{8.75} & 412.6\multirunstd{120} \\
Transformer$^{\dagger}$ & 0.305\multirunstd{0.001} & 0.444\multirunstd{0.001} & 8.3\multirunstd{0.58}  & 32.5\multirunstd{1.6}  & 0.238\multirunstd{0.006} & 0.390\multirunstd{0.012} & 25.0\multirunstd{1.32} & 356.3\multirunstd{35} \\
FEDformer$^{\dagger}$   & 0.315\multirunstd{0.004} & 0.452\multirunstd{0.005} & 9.0\multirunstd{0.00}  & 33.4\multirunstd{0.6}  & 0.231\multirunstd{0.023} & 0.400\multirunstd{0.015} & 30.7\multirunstd{1.15} & 252.0\multirunstd{18} \\
Diffusion-TS$^{\flat}$  & \underline{0.142}\multirunstd{0.007} & \underline{0.251}\multirunstd{0.011} & 30.7\multirunstd{2.08} & 95.6\multirunstd{1.5}  & \underline{0.144}\multirunstd{0.019} & \underline{0.231}\multirunstd{0.026} & 25.0\multirunstd{4.00} & \underline{141.0}\multirunstd{55} \\
TimeLLM$^{\ddagger}$    & 0.395\multirunstd{0.003} & 0.539\multirunstd{0.001} & 13.0\multirunstd{0.00} & \underline{35.1}\multirunstd{0.6} & 0.347\multirunstd{0.004} & 0.452\multirunstd{0.006} & \underline{15.8}\multirunstd{0.58} & 148.5\multirunstd{10} \\
\hline
\rowcolor{icfblue}
\textbf{ICF-DLM (Ours)}$^{\star}$ & \textbf{0.122}\multirunstd{0.003} & \textbf{0.178}\multirunstd{0.004} & \textbf{8.0}\multirunstd{0.15} & \textbf{18.6}\multirunstd{0.5} & \textbf{0.064}\multirunstd{0.002} & \textbf{0.181}\multirunstd{0.003} & \textbf{9.2}\multirunstd{0.17} & \textbf{35.2}\multirunstd{0.8} \\
\hline
\end{tabular}
\end{table*}
\subsubsection{Evaluation Metrics}

 We report four metrics, all lower-the-better. \textbf{MSE / MAE} are squared and absolute errors on the local waveform $\mathbf{w}_{\text{local}}$.  \textbf{PTE} is the Peak Timing Error, $|\hat{t}_{\text{peak}} - t_{\text{peak}}^*|$. \textbf{YE} is the Yield Error, the relative error of $Y_{DT}$ in percent.

\subsection{Benchmark Results}
\label{sec:benchmark}

\subsubsection{Simulation Main Results}
We evaluate on the 5{,}000-shot simulation test set of ICFBench. 
As shown in Table~\ref{tab:results}, ICF-DLM achieves the best MSE, 
MAE, PTE, and YE across all baselines, with MSE $0.122$ and YE 
$18.6\%$. Compared to classical deep time-series models, ICF-DLM 
reduces MSE by $60.0\%$ over the strongest baseline Transformer 
($0.305$) and YE by $39.8\%$ over the strongest baseline GRU
($30.9\%$), likely because the explicit decomposition of yield, peak 
timing, and local waveform captures the nonlinearity of ICF dynamics 
that generic deep sequence models do not target. Compared to 
Diffusion-TS, ICF-DLM reduces MSE by $14.1\%$ ($0.142\!\rightarrow\!0.122$) 
and PTE by $73.9\%$ ($30.7\!\rightarrow\!8.0$ steps), with 
physics-driven rewards and hierarchical decomposition providing an 
additional inductive bias beyond a generic diffusion backbone.
Compared to TimeLLM, ICF-DLM reduces MSE by $69.1\%$ ($0.395\!\rightarrow\!0.122$) 
and PTE by $38.5\%$ ($13.0\!\rightarrow\!8.0$ steps), consistent with 
the picosecond-scale burst window benefiting from bidirectional 
diffusion refinement rather than left-to-right decoding.

% We evaluate on the 5{,}000-shot simulation test set of ICFBench. 
% {As shown in Table~\ref{tab:results}, ICF-DLM achieves the 
% best MSE, MAE, PTE, and YE across all baselines, with MSE $0.059$ 
% and PTE $9.7$ steps. Compared to classical deep time-series models, 
% ICF-DLM reduces MSE by $69.9\%$ over the strongest baseline 
% Transformer ($0.196$) and PTE by $39.4\%$ over FEDformer ($16.0$ 
% steps), likely because the explicit decomposition of yield, peak 
% timing, and local waveform captures the nonlinearity of ICF dynamics 
% that generic deep sequence models do not target. Compared to 
% diffusion-based time-series baselines (CSDI, Diffusion-TS) that 
% report MSE around \todored{X.XXX} and \todored{X.XXX} respectively, 
% ICF-DLM's physics-driven rewards and hierarchical decomposition 
% provide an additional inductive bias beyond a generic diffusion 
% backbone. Compared to LLM-based baselines (TimeLLM, LPI-LLM) that 
% report PTE above $150$ steps under their original recipes, ICF-DLM 
% reduces PTE by $93.6\%$ over LPI-LLM ($151.3$ steps), consistent with 
% the picosecond-scale burst window benefiting from bidirectional 
% diffusion refinement rather than left-to-right decoding.}

\subsubsection{Physical-Experimental Main Results}

All models are pre-trained on simulation data and fine-tuned on the experimental training set.

As shown in Table~\ref{tab:results}, ICF-DLM achieves the best MSE, 
MAE, PTE, and YE on the 50-shot held-out test set. Compared to 
classical deep time-series models, ICF-DLM reduces MSE by $72.3\%$ 
over the strongest baseline FEDformer ($0.231$) and PTE by $39.5\%$
over the strongest baseline RNN ($15.2$ steps). Classical recurrent 
baselines (LSTM, GRU) show notable PTE inflation on experimental 
data ($20.7$ and $25.7$ steps) compared to their simulation 
performance ($10.7$ and $9.0$ steps), while ICF-DLM maintains PTE 
within $9.2$ steps across both distributions, likely due to the 
explicit decomposition of peak timing as a separate prediction 
target. Compared to TimeLLM, ICF-DLM reduces MSE by $81.6\%$ 
($0.347\!\rightarrow\!0.064$) and PTE by $41.8\%$ ($15.8\!\rightarrow\!9.2$ 
steps), with bidirectional diffusion refinement and physics-driven 
rewards transferring cleanly across the sim-to-real gap. The high YE of all baselines on experimental data (above $140\%$) 
reflects the difficulty of sim-to-real transfer 
(Appendix~\ref{appendix:dataset}). ICF-DLM, in contrast, transfers cleanly across the gap.
% As shown in Table~\ref{tab:results}, ICF-DLM achieves the 
% best MSE, PTE, and YE on the 50-shot held-out test set. Specifically, 
% simulation pre-training improves MSE from \todored{X.XXX} to $0.064$, 
% PTE from \todored{XX.X} to $9.2$ steps, and YE from \todored{XX}\% 
% to $35\%$ compared to the version trained from scratch on the 
% 182-shot experimental split alone, confirming that the simulation 
% corpus supplies a transferable inductive bias. Compared to classical 
% recurrent baselines (LSTM, GRU), which show larger PTE on 
% experimental data than on simulation (around $160$ vs.\ $19$ steps), 
% ICF-DLM maintains PTE within $9.2$ steps across both distributions, 
% likely due to the explicit decomposition of peak timing as a separate 
% prediction target. Compared to LLM-based baselines (TimeLLM, LPI-LLM) 
% that report PTE around $160$ steps on experimental data, ICF-DLM's 
% bidirectional diffusion refinement and physics-driven rewards 
% together transfer cleanly across the sim-to-real gap. The lower YE on the experimental set ($35\%$) than on the simulation 
% set ($69\%$) is consistent with the higher mean yield of experimental 
% shots, which enlarges the denominator of the relative-error metric 
% (Appendix~\ref{appendix:dataset}).

\subsubsection{Real-world ICF Experiments Guided by ICF-DLM}
\label{sec:guided}

\begin{figure}[!t]
\centering
\includegraphics[width=\columnwidth]{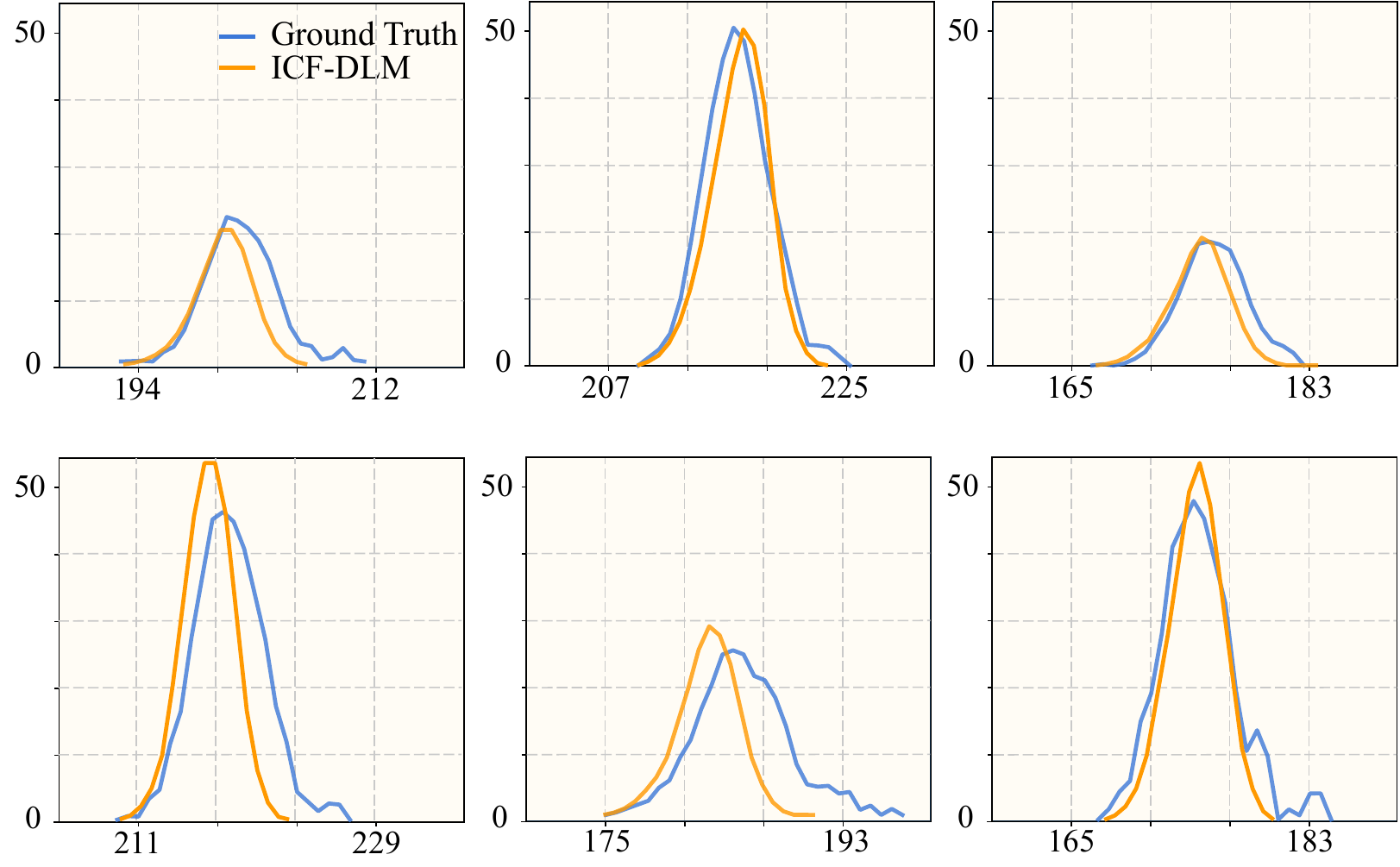}

\caption{Comparison of ICF-DLM predictions (orange solid) versus physical experimental results (blue solid) for six expert-selected input configurations.}
\label{fig:guided_exp}
\vspace{-1em}
\end{figure}

To validate the practical utility of ICF-DLM for guiding real ICF 
experiments, we collaborate with domain experts to conduct a 
prospective validation study. Domain experts propose six input 
configurations of scientific interest for exploring new implosion 
regimes, ICF-DLM predicts the expected neutron-rate waveforms before 
any shot is taken, the corresponding physical experiments are 
subsequently conducted, and predictions are evaluated against the 
measured outcomes.

Figure~\ref{fig:guided_exp} compares ICF-DLM predictions and physical-experimental results. The close agreement between predicted and measured waveforms demonstrates that ICF-DLM can reliably guide experimental design by providing accurate forecasts before costly physical shots are conducted. The agreement on both peak timing and waveform amplitude suggests that the hierarchical decomposition transfers to expert-proposed configurations beyond the random held-out split, offering useful predictive information for experimental planning.

\subsection{Ablation Studies}
\label{sec:ablation}

\paragraph{Effect of Backbone and Training Stage.}
We ablate two design axes. (i) The \emph{backbone}. We replace the
masked-language diffusion model (LLaDA-8B) backbone of ICF-DLM
with an autoregressive model (LLaMA-3-8B) under identical numeric
tokenization, prompt, and LoRA configuration. (ii) The \emph{training
stage}. We remove PPO with the physics-driven rewards and keep only
SFT.
As shown in Table~\ref{tab:ab_backbone}, replacing AR with
a DLM under the same SFT + PPO pipeline lowers MSE from $0.129$ to
$0.064$ and PTE from $11.6$ to $9.2$ steps. 
Removing simulation
pre-training raises MSE from $0.064$ to $0.150$ and PTE from $9.2$
to $33.1$ steps, confirming the value of sim-to-real transfer.
Adding PPO on the DLM
further lowers MSE from $0.100$ to $0.064$ and halves YE from $68\%$
to $35\%$. The DLM gain is consistent with bidirectional denoising
revising yield, peak timing, and local waveform jointly in each step.
The PPO gain reflects physical information about $Y_{DT}$ and
$t_{\text{peak}}$ that the pointwise SFT loss does not provide. 

\begin{table}[h]
\centering
\caption{Backbone $\times$ training stage on the test set.}
\label{tab:ab_backbone}
\vspace{-0.5em}
\resizebox{0.95\columnwidth}{!}{
\setlength\tabcolsep{5pt}
\renewcommand\arraystretch{1.1}
\begin{tabular}{l|ccc}
\hline\thickhline
\rowcolor{mygrey}
\textbf{Variant} & \textbf{MSE} & \textbf{PTE} & \textbf{YE (\%)} \\
\hline
{AR LLaMA-3-8B, SFT + PPO}   & {0.129} & {11.6} & {39} \\

{DLM LLaDA-8B, no simulation pre-train}   & {0.150} & {33.1} & {142} \\

{DLM LLaDA-8B, SFT only}     & 0.100 & 9.6 & 68 \\

{DLM LLaDA-8B, SFT + PPO } & \textbf{{0.064}} & \textbf{9.2} & \textbf{35} \\
\hline
\end{tabular}
}
\vspace{-1em}
\end{table}

\paragraph{Effect of Reward Components.}
We ablate three reward terms that each target a distinct physical
quantity. (i) The waveform reward $r_{\text{wave}}$ supervises the
local waveform shape $\mathbf{w}_{\text{local}}$. (ii) The magnitude
reward $r_{\text{mag}}$ supervises the total neutron yield $Y_{DT}$.
(iii) The peak reward $r_{\text{peak}}$ supervises the bang time
$t_{\text{peak}}$. Each reward is removed in turn while the other two
remain.
{As shown in Table~\ref{tab:ab_reward}, removing $r_{\text{wave}}$
hurts most on MSE (from $0.064$ to $0.360$) and YE (from $35\%$ to
$186\%$). Removing $r_{\text{mag}}$ hurts most on PTE (from $9.2$ to
$17.8$ steps) and YE (from $35\%$ to $148\%$). Removing $r_{\text{peak}}$
inflates PTE (from $9.2$ to $11.9$ steps).} The three rewards are
synergistic, each regularizing the others through shared physical
structure.

\begin{table}[h]
\centering
\caption{Reward component ablation on the test set.}
\label{tab:ab_reward}
\vspace{-0.5em}
\resizebox{0.75\columnwidth}{!}{
\setlength\tabcolsep{5pt}
\renewcommand\arraystretch{1.1}
\begin{tabular}{l|ccc}
\hline\thickhline
\rowcolor{mygrey}
\textbf{Variant} & \textbf{MSE} & \textbf{PTE} & \textbf{YE (\%)} \\
\hline
Full reward  & \textbf{{0.064}} & \textbf{9.2} & \textbf{35} \\
W/o $r_{\text{peak}}$ & 0.087 & 11.9 & 93 \\
W/o $r_{\text{mag}}$  & 0.128 & 17.8 & 148 \\
W/o $r_{\text{wave}}$ & 0.360 & 16.4 & 186 \\
\hline
\end{tabular}
}
\vspace{-1em}
\end{table}

\paragraph{Effect of Remasking Strategy.}
We ablate four remasking strategies that decide which tokens are
re-sampled at each denoising step. (i) \emph{Confidence} re-masks
tokens with the lowest predicted probability. (ii) \emph{Margin
confidence} re-masks tokens with the smallest gap between top-1 and
top-2 probabilities. (iii) \emph{Random} re-masks uniformly at random.
(iv) \emph{Negative entropy} re-masks tokens with the highest
predictive entropy.
{As shown in Table~\ref{tab:ab_remask}, confidence remasking
gives the best numbers (MSE $0.064$, PTE $9.2$, YE $35\%$),
outperforming margin confidence by $0.016$ MSE, random by $0.032$,
and negative entropy by $0.070$.}Confidence remasking concentrates each refinement step on the tokens
the model is least certain about, which matches the iterative
denoising paradigm.

\begin{table}[h]
\centering
\caption{Remasking strategy on the test set.}
\label{tab:ab_remask}
\vspace{-0.5em}
\resizebox{0.75\columnwidth}{!}{
\setlength\tabcolsep{5pt}
\renewcommand\arraystretch{1.1}
\begin{tabular}{l|ccc}
\hline\thickhline
\rowcolor{mygrey}
\textbf{Variant} & \textbf{MSE} & \textbf{PTE} & \textbf{YE (\%)} \\
\hline
Confidence  & \textbf{{0.064}} & \textbf{9.2} & \textbf{35} \\
Margin confidence & 0.080 & 9.8  & 38 \\
Random            & 0.096 & 10.4 & 43 \\
Neg.\ entropy     & 0.134 & 11.7 & 45 \\
\hline
\end{tabular}
}
\vspace{-1em}
\end{table}

\paragraph{Effect of Block Size.}
The block size controls how many tokens are denoised jointly before
the next block, where larger blocks model longer-range dependency in
one shot.
{As shown in Table~\ref{tab:ab_block}, block size $128$ gives
the best numbers (MSE $0.064$, PTE $9.2$, YE $35\%$), outperforming
$64$ by $0.026$ MSE and $32$ by $0.018$ MSE.} Larger blocks improve
global coherence across the joint generation of $Y_{DT}$,
$t_{\text{peak}}$, and $\mathbf{w}_{\text{local}}$.

\begin{table}[h]
\centering
\caption{Block size on the test set.}
\label{tab:ab_block}
\vspace{-0.5em}
\resizebox{0.7\columnwidth}{!}{
\setlength\tabcolsep{5pt}
\renewcommand\arraystretch{1.1}
\begin{tabular}{l|ccc}
\hline\thickhline
\rowcolor{mygrey}
\textbf{Variant} & \textbf{MSE} & \textbf{PTE} & \textbf{YE (\%)} \\
\hline
128  & \textbf{{0.064}} & \textbf{9.2} & \textbf{35} \\
64         & 0.090 & 9.6 & 37 \\
32         & 0.082 & 9.9 & 38 \\
\hline
\end{tabular}
}
\vspace{-1em}
\end{table}

\paragraph{Effect of Denoise Steps.}
The number of denoise steps controls how many refinement iterations
the diffusion decoder takes at inference, where more steps allow 
finer token correction at the cost of compute.
{As shown in Table~\ref{tab:ab_steps}, $128$ steps gives the
best numbers (MSE $0.064$, PTE $9.2$, YE $35\%$), outperforming $64$
steps by $0.023$ MSE and $32$ steps by $0.032$ MSE.} More steps allow 
ICF-DLM to perform additional bidirectional refinement 
iterations over the masked numeric tokens, progressively improving 
prediction accuracy, so we use $128$ steps as the default.
\begin{table}[h]
\centering
\caption{Denoise steps on the test set.}
\label{tab:ab_steps}
\vspace{-0.5em}
\resizebox{0.7\columnwidth}{!}{
\setlength\tabcolsep{5pt}
\renewcommand\arraystretch{1.1}
\begin{tabular}{l|ccc}
\hline\thickhline
\rowcolor{mygrey}
\textbf{Variant} & \textbf{MSE} & \textbf{PTE} & \textbf{YE (\%)} \\
\hline
Steps = 32        & 0.096 & 9.8 & 39 \\
Steps = 64       & 0.087 & 9.4 & 37 \\
Steps = 128  & \textbf{{0.064}} & \textbf{9.2} & \textbf{35} \\
\hline
\end{tabular}
}
\vspace{-1em}
\end{table}

\paragraph{Effect of Architectural Choices.}
We further ablate three architectural choices of ICF-DLM by replacing
each component while holding the rest of the pipeline fixed. (i) The
\emph{LM backbone}, replaced by a conditional MLP regressor that
consumes the same $(T, P)$ input and predicts $(Y_{DT}, t_{\text{peak}},
\mathbf{w}_{\text{full}})$ through dedicated regression heads. (ii) The
\emph{hierarchical decomposition}, replaced by a flat variant that
predicts the full 512-step waveform directly without the
$Y_{DT}\!\rightarrow\!t_{\text{peak}}\!\rightarrow\!\mathbf{w}_{\text{local}}$
factorization. (iii) The \emph{numeric tokenization}, kept in place but
read out with expected-value (``soft'') decoding $\sum_i p_i\,v_i$ at
the $\langle l\rangle$ and $\langle t\rangle$ positions instead of
argmax, which approximates a continuous regression head reading off
the same hidden state.
As shown in Table~\ref{tab:ab_arch}, removing the
decomposition is the most damaging change, raising MSE from $0.064$
to $0.251$ (a $3.9\times$ increase) and YE to nearly $100\%$ as the
model collapses to a near-constant baseline output. Removing the LM
backbone preserves peak timing (PTE $9.4$ vs. ICF-DLM's $9.2$) but
inflates YE from $35\%$ to $55\%$, with the gap driven by the
low-yield tail ($Y_{DT}{<}1.5$) on which the MLP is unstable.
Switching argmax to soft decoding leaves three metrics largely
unchanged (PTE $13.0$, YE $45\%$). Hierarchical decomposition
contributes the largest gain among the three. The LM backbone narrows
the right tail of yield error under simulation-to-experiment shift,
while the discrete numeric tokenization is a representational choice
rather than a performance constraint.

\begin{table}[h]
\centering
\caption{Architectural choice ablations on the test set.}
\label{tab:ab_arch}
\vspace{-0.5em}
\resizebox{0.95\columnwidth}{!}{
\setlength\tabcolsep{5pt}
\renewcommand\arraystretch{1.1}
\begin{tabular}{l|ccc}
\hline\thickhline
\rowcolor{mygrey}
\textbf{Variant} & \textbf{MSE} & \textbf{PTE} & \textbf{YE (\%)} \\
\hline
{W/o decomposition}         & {0.251} & {179.5} & {99} \\
{W/o numeric tokenization}  & {0.064} & {13.0}  & {45}   \\
{W/o LM backbone}           & {0.146} & {9.4}   & {55} \\
\hline
\end{tabular}
}
\vspace{-1em}
\end{table}

\section{Discussion}
ICF is a natural stress test for language-model-based numeric prediction. Three observations follow that should generalize beyond fusion.
\indent\textbf{\textit{Bidirectional decoding helps when outputs are globally coupled.}}
Under matched conditions, the diffusion backbone improves over a matched-AR LLaMA-3-8B on waveform MSE with smaller gains on peak-timing and yield error (Table~\ref{tab:ab_backbone}). Iterative denoising for time-series has been explored by CSDI \cite{tashiro2021csdi} and Diffusion-TS \cite{yuan2024diffusionts}, but those models operate on continuous values with regression backbones and assume input-output modality match; the structural advantage we observe here is specifically for LM-style discrete-token decoding under exogenous-driven prediction, where the model has no temporal scaffolding from past observations. The qualitative reason is that AR decoding commits to a peak location early, and the surrounding waveform must remain consistent with that commitment, whereas bidirectional denoising lets the peak adjust as the waveform crystallizes. We further verify this by tracing the denoising trajectory on the test set, where shots undergo bidirectional revision during decoding, reducing PTE by $2.48$ steps (Appendix~\ref{appendix:trajectory}). We expect this advantage to recur for tasks whose outputs factorize into globally coupled physical quantities that resist a natural left-to-right ordering.\\
\indent\textbf{\textit{Numeric tokenization needs metric-aware supervision.}}
Discrete numeric tokenization induces a quantization error well below the typical prediction error, so quantization itself is not the bottleneck. The mismatch surfaces during RL. Cross-entropy treats adjacent numeric tokens as fully distinct classes, while the physics-driven rewards see them as nearly equivalent; the magnitude reward $r_{\text{mag}}$ absorbs this mismatch by re-injecting metric structure across the vocabulary. Consistent with this view, removing $r_{\text{mag}}$ degrades MSE and yield error more than removing $r_{\text{peak}}$ (Table~\ref{tab:ab_reward}), even though $r_{\text{peak}}$ is the only reward that directly supervises peak timing.\\
\indent\textbf{\textit{A recipe for LM-based scientific prediction in low-data regimes.}}
Three design choices appear to matter for LM-based numeric prediction when scientific data are scarce and critical events are sparse. First, factorize the output along physically meaningful axes so the model is supervised on the descriptors physicists already use. Second, generate bidirectionally so commitment to globally coupled landmarks can be deferred. Third, pair token cross-entropy with at least one metric-aware reward to close the discretization gap. ICF stresses all three simultaneously; we expect the recipe to apply to other sparse-event scientific prediction tasks, an extension we leave to future work.
\vspace{-0.5em}
\section{Conclusion}
\vspace{-0.5em}
We presented \textsc{ICF-DLM}, to our knowledge the first language-model predictor for ICF neutron-rate waveforms, combining physics-typed output decomposition, bidirectional denoising that defers commitment to peak location, and a physics-driven PPO reward that restores metric structure across numeric tokens. On the held-out experimental test set, it outperforms classical sequence models, LLM-based time-series predictors, and a matched autoregressive LLaMA-3-8B, pointing to a transferable recipe for LM-based scientific prediction in low-data sparse-event regimes.

\section*{Limitations}

Three caveats bound the scope of our claims. First, all experimental data come from a single ICF facility and a single diagnostic, the neutron-rate waveform; generalization to other diagnostics (X-ray imaging, neutron time-of-flight) and to facilities with different driver geometries is left to future work. Second, although the recipe conceptually applies to other scientific time-series with similar structural properties, we have not empirically validated it on domains such as tokamak disruption prediction, supernova light curves, or shock-induced phase transitions. Third, diffusion generation requires multiple forward passes per prediction, so per-sample latency exceeds single-pass predictors. This is acceptable for ICF planning, where each physical shot is far more expensive than inference, but would need distillation~\cite{han2024amd} or few-step variants for real-time settings.

\section*{Ethical Considerations}

This work targets scientific discovery in fusion energy research, a domain with significant potential societal benefit for clean energy generation. The ICFBench data are derived from publicly funded research at national laboratories. Our methods produce fast surrogate predictions of neutron-rate waveforms; they are not a replacement for facility-grade simulation or experimental design tools, and any practical use should retain expert review and standard quality control on the underlying physical predictions.

Inertial confinement fusion overlaps with restricted nuclear-weapons-relevant physics, and predictive surrogates of neutron yield could in principle reduce the experimental cost of designing high-yield implosions for non-peaceful applications. Our contribution is intentionally narrow, namely a tokenizer-level diffusion-LM recipe for structured numeric prediction in low-data regimes, evaluated on simulation-derived data and a small set of facility-released experimental shots. Inference latency, evaluation cost, and the small experimental shot count are explicitly low enough that the artifact does not measurably accelerate weapons-relevant design optimization beyond what existing surrogate physics tools already provide.

\section*{Acknowledgements}

This research was supported by the DOE Fusion Energy Science Program under grant DE-SC0024381. We also gratefully acknowledge the support of the Purdue Polytechnic Institute.

\bibliography{reference_google_scholar}

\clearpage
\appendix
\renewcommand{\thesection}{S\arabic{section}}
\renewcommand{\thetable}{S\arabic{table}}
\renewcommand{\thefigure}{S\arabic{figure}}
\setcounter{section}{0}
\setcounter{table}{0}
\setcounter{figure}{0}

\centerline{\textbf{APPENDIX}}
\vspace{0.5em}
This appendix contains additional details for the EMNLP 2026 submission, titled \textit{``Decomposition-Guided Diffusion Language Models for Inertial Confinement Fusion Prediction''}. The appendix is organized as follows:

\begin{itemize}[leftmargin=*]
\setlength\itemsep{-0.3mm}
\item \S\ref{appendix:dataset} provides detailed descriptions of the ICFBench dataset, including data collection, preprocessing, and statistics.
\item \S\ref{appendix:implementation} extends the Methods formalism with the full training algorithm, hyperparameters, LoRA configuration, prompt template, and inference settings.
\item \S\ref{appendix:baseline} describes baseline implementations.
\item \S\ref{appendix:trajectory} analyzes denoising trajectories to characterize how bidirectional revision contributes to peak placement.
\item \S\ref{appendix:yield_strat} stratifies per-shot errors by ground-truth $Y_{DT}$ magnitude across yield bins.
\item \S\ref{appendix:case_studies} presents qualitative case studies with visualization of predictions.
\item \S\ref{appendix:physical_interpretation} discusses the physical interpretation of hierarchical decomposition.
\item \S\ref{appendix:licenses} lists artifact licenses.
\end{itemize}

%==============================================================================
\section{Dataset Details}\label{appendix:dataset}
%==============================================================================

\subsection{ICFBench Dataset Overview}

The ICFBench dataset is constructed from two complementary sources to enable both large-scale training and realistic evaluation:

\paragraph{Simulation Data}
We utilize 50,000 samples generated from radiation-hydrodynamics (rad-hydro) simulation codes. These simulations span diverse implosion regimes by systematically varying laser pulse shapes, target specifications, and experimental configurations.
Each simulation provides:
\begin{itemize}
    \item \textbf{Input}: 5 scalar target parameters ($\vec{T}$) governing target geometry and a 256-dimensional pulse shape ($\vec{P}$).
    \item \textbf{Output}: Full neutron rate ($\dot{n}$) time-series of 512 steps, total yield $Y_{DT}$, and peak timing $t_{\text{peak}}$.
\end{itemize}

\paragraph{Experimental Data}
{The experimental subset consists of 232 physical shots from 
real ICF experiments, partitioned into 182 training shots and 50 
held-out test shots.}

\subsection{Target Parameter Ranges}

Table~\ref{tab:param_ranges} summarizes the ranges of target parameters ($\vec{T}$) in both simulation and experimental datasets.

\begin{table}[h]
\centering
\caption{Target design parameter ranges for simulation and experimental data.}
\label{tab:param_ranges}
\vspace{-0.5em}
\resizebox{0.95\columnwidth}{!}{
\begin{tabular}{l|cc|cc}
\hline\thickhline
\rowcolor{mygrey}
 & \multicolumn{2}{c|}{\textbf{Simulation}} & \multicolumn{2}{c}{\textbf{Experimental}} \\
\rowcolor{mygrey}
\textbf{Param} & \textbf{Range} & \textbf{Mean} & \textbf{Range} & \textbf{Mean} \\
\hline\hline
$T_0$ & [0.42, 0.76] & 0.64 & [0.51, 0.73] & 0.62 \\
$T_1$ & [0.60, 2.97] & 1.47 & [0.88, 2.90] & 1.52 \\
$T_2$ & [0.08, 0.90] & 0.42 & [0.34, 0.68] & 0.56 \\
$T_3$ & [1.05, 1.47] & 1.26 & [1.04, 1.49] & 1.28 \\
$T_4$ & [0.00, 3.90] & 1.95 & [0.25, 4.62] & 0.57 \\
\hline
\end{tabular}
}
\end{table}

\subsection{Yield Distribution}

The neutron yield $Y_{DT}$ (in units of $10^{19}$) exhibits different distributions between simulation and experimental data:

\begin{itemize}
    \item \textbf{Simulation}: Range $[7.2 \times 10^{-4}, 27.0]$, mean $1.75 \pm 1.82$, median $1.13$
    \item \textbf{Experimental}: Range $[0.16, 17.5]$, mean $4.19$, reflecting selection bias toward higher-yield shots in physical experiments
\end{itemize}

\subsection{Dataset Statistics Summary}

Table~\ref{tab:dataset_stats} summarizes key statistics of the ICFBench dataset.

\begin{table}[h]
\centering
\caption{ICFBench dataset statistics.}
\label{tab:dataset_stats}
\vspace{-0.5em}
\resizebox{0.95\columnwidth}{!}{
\begin{tabular}{l|cc}
\hline\thickhline
\rowcolor{mygrey}
\textbf{Statistic} & \textbf{Simulation} & \textbf{Experimental} \\
\hline\hline
Total samples & 50,000 & 232  \\
Training set & 40,000 & 182  \\
Validation set & 5,000 & --- \\
Test set & 5,000 & 50\\
\hline
Pulse shape & 256 & 256 \\
Target parameters  & 5 & 5 \\
Output length  & 512 & 512 \\
$Y_{DT}$ range & [$7.2 \times 10^{-4}$, 27.0] & [0.16, 17.5] \\
$Y_{DT}$ mean & 1.75 & 4.19 \\
\hline
\end{tabular}
}
\end{table}

%==============================================================================
\section{Implementation Details}\label{appendix:implementation}
%==============================================================================

\subsection{Pipeline Components}\label{appendix:components}

\textsc{ICF-DLM} has three components (Fig.~\ref{fig:overview}). The \textit{input encoder} serializes $\vec{P}$ and $\vec{T}$ into a structured prompt, with continuous values discretized into numeric tokens. The \textit{LLaDA-8B backbone} \cite{nie2025llada} denoises $(Y_{DT}, t_{\text{peak}}, \mathbf{w}_{\text{local}})$ jointly under bidirectional attention. The \textit{output decoder} embeds $\hat{\mathbf{w}}_{\text{local}}$ at $\hat{t}_{\text{peak}}$ within a 512-step trajectory, filling the remainder with a near-zero baseline.

\subsection{Data Preprocessing and Output Reconstruction}\label{appendix:preprocessing}

All input features and output waveforms are min-max normalized to $[-0.5, 0.5]$. Following the numeric-tokenization recipe of ChatTime~\cite{wang2025chattime}, continuous values are discretized into a vocabulary of $10{,}000$ numeric tokens, enabling the diffusion LM to perform regression through classification over the token vocabulary. For the hierarchical decomposition, we extract a local waveform of length $T_{\text{local}}{=}19$ centered at the ground-truth peak time during training; $T_{\text{local}}{=}19$ matches the physical neutron-burst full width in our dataset, so $\mathbf{w}_{\text{local}}$ covers the entire active region while the surrounding $493$ steps carry negligible signal. At inference, the predicted local waveform is embedded into a $512$-step sequence at the predicted peak position, with remaining positions filled by a near-zero baseline ($\epsilon{\approx}10^{-6}$).

\subsection{Training Algorithm}

Algorithm~\ref{alg:icf-dlm} summarizes the overall training procedure of ICF-DLM.

\begin{algorithm}[!t]
\caption{ICF-DLM Training Procedure}
\label{alg:icf-dlm}
\begin{algorithmic}[1]
\REQUIRE Training data $\mathcal{D} = \{(\vec{T}_i, \vec{P}_i, \mathbf{y}_i)\}_{i=1}^N$, denoising steps $S$, learning rate $\eta$
\ENSURE Trained model parameters $\theta$
\STATE Initialize LLaDA backbone with LoRA adapters
\STATE \textbf{// Stage 1: Supervised Fine-Tuning}
\FOR{each epoch in SFT stage}
    \FOR{each batch $(\vec{T}, \vec{P}, \mathbf{y}) \in \mathcal{D}$}
        \STATE Decompose target: $\mathbf{y} \rightarrow (Y_{DT},\; t_{\text{peak}},\; \mathbf{w}_{\text{local}})$
        \STATE Tokenize and construct structured prompt
        \STATE Sample mask ratio $t \sim \text{Uniform}(0, 1)$
        \STATE Forward process: $\mathbf{x}_t = \text{Mask}(\mathbf{x}_0, t)$ \hfill $\triangleright$ mask $t$-fraction of tokens
        \STATE Predict masked tokens: $\hat{\mathbf{x}}_0 = f_\theta(\mathbf{x}_t, \vec{T}, \vec{P})$
        \STATE $\mathcal{L}_{\text{SFT}} = -\sum_{i \in \mathcal{M}} \log p_\theta(x_0^{(i)} \mid \mathbf{x}_t)$
        \STATE Update $\theta \leftarrow \theta - \eta \nabla_\theta \mathcal{L}_{\text{SFT}}$
    \ENDFOR
\ENDFOR
\STATE \textbf{// Stage 2: RL with Physics-Driven Rewards}
\FOR{each epoch in RL stage}
    \FOR{each batch $(\vec{T}, \vec{P}, \mathbf{y}) \in \mathcal{D}$}
        \STATE Generate $(\hat{Y}_{DT}, \hat{t}_{\text{peak}}, \hat{\mathbf{w}}_{\text{local}})$ via $S$-step denoising
        \STATE $r_{\text{mag}} \leftarrow \frac{1}{1+|\hat{Y}_{DT} - Y_{DT}^*|/Y_{DT}^*}$
        \STATE $r_{\text{peak}} \leftarrow \exp(-|\hat{t}_{\text{peak}} - t^*|)$
        \STATE $r_{\text{wave}} \leftarrow -\log(1 + \|\hat{\mathbf{w}} - \mathbf{w}^*\|_2^2)$
        \STATE $r = r_{\text{mag}} + r_{\text{peak}} + r_{\text{wave}} + r_{\text{constraint}}$
        \STATE Update $\theta$ via PPO with reward $r$
    \ENDFOR
\ENDFOR
\RETURN $\theta$
\end{algorithmic}
\end{algorithm}

\subsection{Model and Training Configuration}

We implement ICF-DLM with LLaDA-8B~\cite{nie2025llada} as the bidirectional Transformer backbone (hidden dim 4096, 32 layers, 32 attention heads). The vocabulary extends the base LLaDA vocabulary of 128{,}256 with 10{,}006 additional numeric and special tokens for waveform discretization. We fine-tune with Low-Rank Adaptation (LoRA)~\cite{hu2021lora} at rank $r{=}16$, $\alpha{=}64$, and dropout $0.05$, applied to the q/k/v/o and gate/up/down projections (\textasciitilde$42$M trainable, $0.5\%$ of the full model).

Supervised fine-tuning runs for 5 epochs on $40{,}000$ samples with batch size $32$, learning rate $2{\times}10^{-4}$ under a cosine schedule with $3\%$ warmup, AdamW optimizer ($\beta_1{=}0.9$, $\beta_2{=}0.999$, weight decay $0.01$), and max sequence length $512$. PPO follows for $200$ iterations with batch size $8$, learning rate $1{\times}10^{-5}$, KL coefficient $\beta{=}0.01$, clip ratio $\epsilon{=}0.2$, batch-level advantage normalization, and reward clipping to $[-3, 3]$. At inference, we use $128$ denoising steps with confidence-based unmasking and greedy decoding ($\tau{=}0$).

\subsection{Prompt Template}

Table~\ref{tab:prompt_template} illustrates the prompt format used for training and inference. All continuous values are normalized to $[-0.5, 0.5]$ and discretized into 10,000 tokens.

\begin{table}[h]
\centering
\caption{Prompt template for ICF-DLM. Values are normalized and tokenized using the \texttt{\#\#\#value\#\#\#} format.}
\label{tab:prompt_template}
\vspace{-0.5em}
\resizebox{\columnwidth}{!}{
\begin{tabular}{p{0.25\columnwidth}|p{0.75\columnwidth}}
\hline\thickhline
\rowcolor{mygrey}
\textbf{Component} & \textbf{Format} \\
\hline\hline
\multicolumn{2}{l}{\cellcolor{icfblue}\textit{\textbf{Input Prompt}}} \\
\hline
Task & \texttt{Predict L (yield), T (peak), W (waveform)} \\
\hline
Target Params & $T_0$=0.644, $T_1$=1.445, $T_2$=0.564, $T_3$=1.360, $T_4$=0.455 \\

\hline
Pulse Shape & \texttt{\#\#\#-0.4961\#\#\# \#\#\#-0.4903\#\#\# \#\#\#-0.4961\#\#\# ...} {\scriptsize(256 tokens)} \\
\hline
\end{tabular}
}
\end{table}

%=============================================================================
\section{Baseline Implementations}\label{appendix:baseline}

\noindent\textbf{Classical baselines.} We implement RNN, LSTM, and GRU as 2-layer bidirectional networks with hidden size $256$ and dropout $0.1$. The Transformer baseline is a standard encoder--decoder with hidden dim $512$, 8 attention heads, 6 encoder/decoder layers, and feedforward dim $2048$. FEDformer uses the official implementation with default hyperparameters~\cite{zhou2022fedformer}.

\noindent\textbf{LLM-based baselines.} TimeLLM~\cite{jin2024timellm} uses LLaMA-3-8B with a reprogramming layer for time-series adaptation, following the original recipe.

% \subsection{Evaluation Protocol}

% \todored{All baselines and ICF-DLM are trained on the same data: 40{,}000 simulation samples under a waveform reconstruction objective, together with 182 cleaned experimental training shots. The simulation columns of Table~\ref{tab:results} are evaluated on the 5{,}000-shot simulation test split, and the physical-experimental columns on the 50-shot held-out test set (see Appendix~\ref{appendix:dataset}). To ensure fair comparison across methods we adopt the following unified protocol:

% \item \emph{Splits.} All models share the same train, validation, and test splits on both data sources.
% \item \emph{Peak timing.} The PTE reported is the full-waveform argmax of the reconstructed 512-step trajectory, $\hat{t}_{\text{peak}} = \argmax_t \hat{y}_t$, computed identically for every method. ICF-DLM's internal $\hat{t}_{\text{peak}}$ scalar head is used only to position the emitted 19-step local waveform within the 512-step reconstruction; it is not the reported estimator.
% \item \emph{Yield.} The yield reported in Table~\ref{tab:results} is $\sum_t \hat{y}_t$ for baselines and the explicit $\hat{Y}_{DT}$ head for ICF-DLM.

%==============================================================================

{\section{Denoising-Trajectory Analysis}\label{appendix:trajectory}
The matched-AR comparison in the main text shows ICF-DLM improving 
over an AR backbone. We examine whether the diffusion backbone uses 
bidirectional revision during decoding. We instrument greedy 
decoding on all $N{=}50$ test shots and log, at every 
step, the argmax over numeric tokens at the $\langle t\rangle$ 
position. Table~\ref{tab:trajectory} reports the trajectory 
statistics. $60\%$ of test shots show some peak revision, $44\%$ 
show $\geq 10$-step revision, and $34\%$ show $\geq 20$-step 
revision, with revisions contributing an average PTE reduction of 
$2.48$ steps over the denoising trajectory. A control that forces 
$\langle t\rangle$ to commit at step $0$ and freezes it for the 
remaining $127$ steps confirms this gap, consistent with 
bidirectional revision contributing to peak placement.}

\begin{table}[h]
\centering
\caption{Denoising-trajectory statistics on the test set ($N{=}50$). 
See \S\ref{appendix:trajectory} for the protocol.}
\label{tab:trajectory}
\fontsize{8}{10}\selectfont
\setlength\tabcolsep{4pt}
\begin{tabular}{l|c}
\hline\thickhline
\rowcolor{mygrey}
\textbf{Statistic} & \textbf{Value} \\
\hline
Shots with any peak revision (span $\geq 1$)       & 30 / 50 (60\%) \\
Shots with significant revision (span $\geq 10$)    & 22 / 50 (44\%) \\
Shots with strong revision (span $\geq 20$)        & 17 / 50 (34\%) \\
\hline
PTE reduction across the denoising trajectory      & $-2.48$ steps \\
PTE increase under the frozen-$\langle t\rangle$ control & $+2.48$ steps \\
\hline
Revisions that move $\langle t\rangle$ toward GT   & 68\% \\
Mean PTE improvement on revised shots              & $-6.53$ steps \\
\hline
\end{tabular}
\end{table}

\section{Yield-Regime Stratification}\label{appendix:yield_strat}
We stratify per-shot errors on the held-out test set by ground-truth 
$Y_{DT}$ magnitude (Table~\ref{tab:yield_strat}). PTE is stable 
across the three yield bins (low, mid, high), indicating that peak 
localization does not degrade for low-yield shots. YE decreases monotonically from the low-yield bin to the high-yield 
bin, reflecting that the denominator of the relative-error metric 
grows with yield magnitude.
\begin{table}[h]
\centering
\caption{Yield-stratified per-shot error on the test set.}
\label{tab:yield_strat}
\fontsize{8}{10}\selectfont
\setlength\tabcolsep{6pt}
\begin{tabular}{l|c|ccc}
\hline\thickhline
\rowcolor{mygrey}
\textbf{Bin} & \textbf{$n$} & \textbf{PTE} & \textbf{YE (\%)} & \textbf{MSE} \\
\hline
overall                          & 50 & 9.2  & 35.1 & 0.064 \\
low ($Y_{DT}{<}1.5$)             & 8  & 7.1  & 64.9 & 0.184 \\
mid ($1.5{\le}Y_{DT}{<}5$)       & 23 & 10.7 & 33.6 & 0.043 \\
high ($Y_{DT}{\ge}5$)            & 19 & 8.2  & 24.5 & 0.040 \\
\hline
\end{tabular}
\end{table}

\subsection{Per-Component Reward Analysis}

Table~\ref{tab:reward_breakdown} shows the breakdown of reward components during RL training, demonstrating that all three objectives improve over training.

\begin{table}[h]
\centering
\caption{Reward component breakdown (mean $\pm$ std over last 50 training steps).}
\label{tab:reward_breakdown}
\vspace{-0.5em}
\resizebox{0.95\columnwidth}{!}{
\begin{tabular}{l|ccc}
\hline\thickhline
\rowcolor{mygrey}
\textbf{Stage} & \textbf{$r_{\text{peak}}$} & \textbf{$r_{\text{mag}}$} & \textbf{$r_{\text{wave}}$} \\
\hline\hline
Initial (SFT) & $0.42 \pm 0.15$ & $0.38 \pm 0.12$ & $-1.24 \pm 0.31$ \\
After RL & $0.68 \pm 0.11$ & $0.56 \pm 0.09$ & $-0.72 \pm 0.18$ \\
\hline
\end{tabular}
}
\end{table}

\subsection{Comparison with $Y_{DT}$ Surrogate}
Under the same SFT-only training setting (40,000 samples, 5 epochs), a dedicated $Y_{DT}$ surrogate model that only predicts the scalar energy descriptor achieves 83.6\% $Y_{DT}$ yield error. In contrast, our method under the same SFT-only setting not only produces full waveform predictions but also achieves better $Y_{DT}$ accuracy (68\% YE, cf.\ \texttt{SFT only} row in Table~\ref{tab:ab_backbone}) as a byproduct of the hierarchical decomposition.

\section{Case Studies}\label{appendix:case_studies}

\begin{figure*}[!t]
\centering
\includegraphics[width=0.9\textwidth]{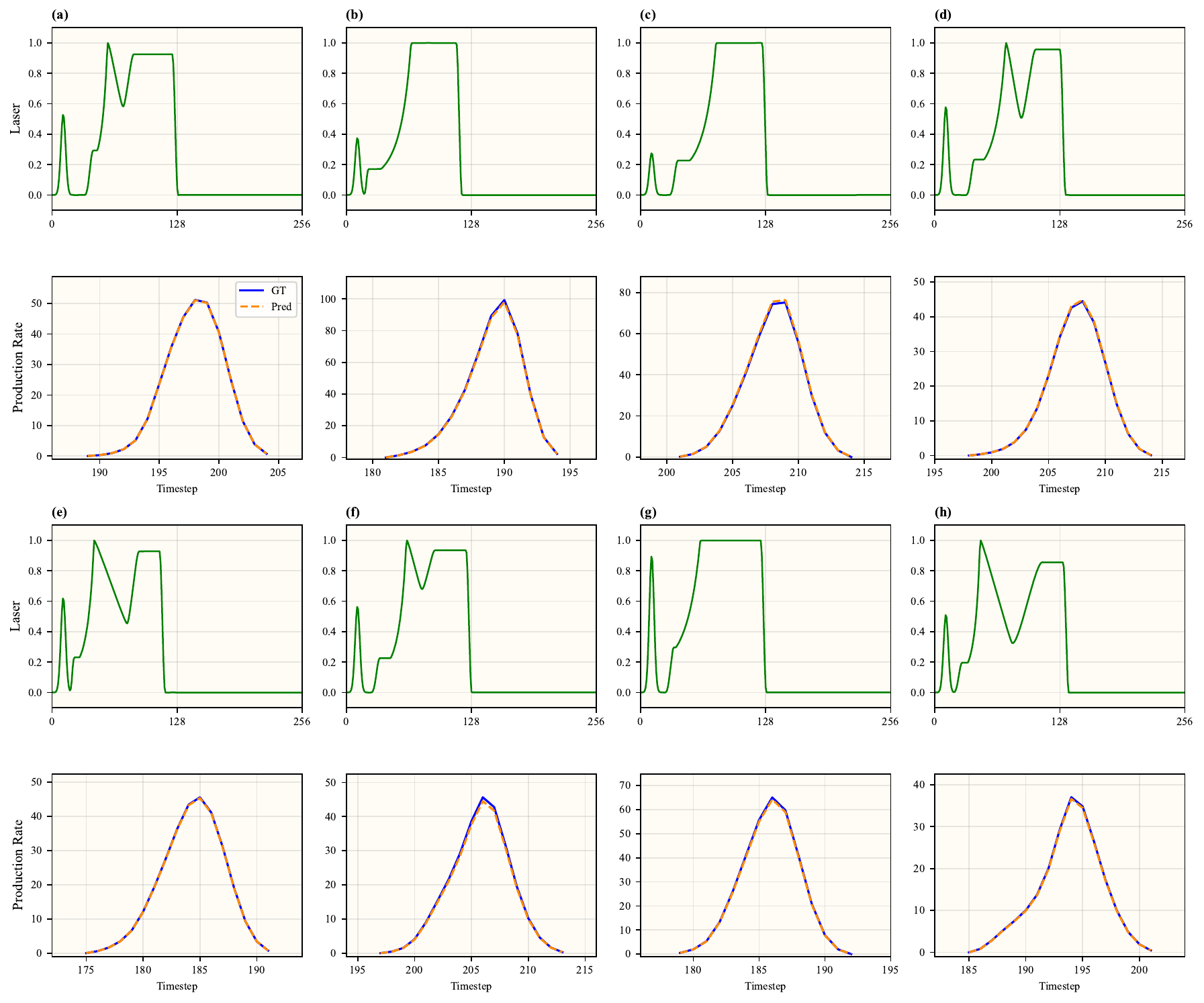}
\caption{Successful prediction cases on simulation data. Top rows show laser input waveforms; bottom rows show $Y_{DT}$ output with ground truth (blue) and ICF-DLM prediction (orange dashed). Labels indicate $Y_{DT}$ values (GT$\rightarrow$Pred) and timing error ($\Delta t$).}
\label{fig:success_cases}
\end{figure*}

\subsection{Successful Predictions}

Figure~\ref{fig:success_cases} shows eight representative cases spanning
$Y_{DT}$ values from $2.6$ to $5.7$ under diverse pulse shapes. Predicted
$Y_{DT}$ deviates from ground truth by below $3\%$ and peak timing aligns
with $\Delta t{=}0$ across all cases, indicating that the hierarchical
decomposition generalizes across pulse-shape regimes.

\subsection{Failure Mode Analysis}

Prediction quality degrades in the tails of the yield distribution.
Shots with $Y_{DT} < 0.5$ or $Y_{DT} > 5.0$ carry the largest relative
error. Table~\ref{tab:yield_strat} shows the same trend across yield
bins. Both tails are thinly covered by the training data. Targeted
augmentation or importance-weighted training may close this gap.

%==============================================================================

\section{Physical Interpretation}\label{appendix:physical_interpretation}
%==============================================================================

The hierarchical decomposition in ICF-DLM mirrors the workflow employed by ICF physicists when analyzing radiation-hydrodynamics simulation outputs. Domain experts typically begin by examining total neutron yield to assess overall implosion performance. They then identify bang time to understand compression dynamics and hot-spot formation. Finally, they analyze the detailed burn rate profile around peak emission to diagnose burn propagation and confinement. By encoding this expert workflow into the model architecture, we provide strong inductive biases that improve both prediction accuracy and physical interpretability.

\subsection{Learned Representations}

Analysis of attention patterns reveals that the model learns physically meaningful relationships. Peak timing prediction attends strongly to laser pulse rise time and peak power, consistent with the known physics of shock timing in ICF implosions. Yield prediction correlates with integrated laser energy and target coupling efficiency indicators, reflecting the energy balance considerations central to ICF performance. Local waveform generation shows strong dependence on the predicted $Y_{DT}$ and $t_{\text{peak}}$ values, confirming that the hierarchical conditioning mechanism effectively propagates information across prediction stages.
\section{Artifact Licenses}\label{appendix:licenses}

LLaDA-8B \cite{nie2025llada} is licensed under \href{https://opensource.org/license/mit/}{MIT};
LLaMA-3-8B \cite{llama3herd2024} is licensed under \href{https://llama.meta.com/llama3/license}{Llama 3 Community License Agreement};
LoRA \cite{hu2021lora}, PPO \cite{schulman2017ppo}, and the baseline implementations
\cite{zhou2022fedformer,yuan2024diffusionts,jin2024timellm}
are licensed under \href{https://www.apache.org/licenses/LICENSE-2.0}{Apache-2.0} or \href{https://opensource.org/license/mit/}{MIT}.
Our use is consistent with their intended research purposes.

\end{document}